\documentclass{article}

\usepackage[numbers]{natbib}
\usepackage[preprint]{score_workshop}

\usepackage[T1]{fontenc}
\usepackage[utf8]{inputenc}
\usepackage{amsmath}
\usepackage{amsfonts}
\usepackage{bm}
\usepackage{booktabs}
\usepackage{dcolumn}
\usepackage{graphicx}
\usepackage{hyperref}
\usepackage{nicefrac}
\usepackage{xcolor}
\usepackage{xspace}
\usepackage{caption}
\usepackage{subcaption}
\usepackage[export]{adjustbox}

\graphicspath{{./figures/}{./}}

\hypersetup{
    colorlinks,
    linkcolor={red!50!black},
    citecolor={blue!50!black},
    urlcolor={blue!80!black}
}

\newcommand{\smuse}[1][]{s^{\rm \scriptscriptstyle \MUSE}\ifthenelse{\equal{#1}{}}{}{_#1}}
\newcommand{\gmuse}[1][]{g^{\rm \scriptscriptstyle \MUSE}\ifthenelse{\equal{#1}{}}{}{_#1}}
\newcommand{\hmuse}[1][]{h^{\rm \scriptscriptstyle \MUSE}\ifthenelse{\equal{#1}{}}{}{_{#1}}}
\newcommand{\smap}[1][]{s^{\rm \scriptscriptstyle MAP}\ifthenelse{\equal{#1}{}}{}{_#1}}

\newcommand{\x}{x}
\newcommand{\MUSE}{MUSE\xspace}

\begin{document}

\title{Improved Marginal Unbiased Score Expansion (MUSE) via Implicit Differentiation}

\author{%
  Marius Millea\\
  Department of Physics, University of California, Berkeley, CA 94720, USA\\
  Department of Physics, University of California, Davis, CA 95616, USA\\
  \texttt{mariusmillea@gmail.com}
}

\maketitle

\begin{abstract}
We apply the technique of implicit differentiation to boost performance, reduce numerical error, and remove required user-tuning in the Marginal Unbiased Score Expansion (MUSE) algorithm for hierarchical Bayesian inference. We demonstrate these improvements on three representative inference problems: 1) an extended Neal's funnel 2) Bayesian neural networks, and 3) probabilistic principal component analysis. On our particular test cases, MUSE with implicit differentiation is faster than Hamiltonian Monte Carlo by factors of 155, 397, and 5, respectively, or factors of 65, 278, and 1 without implicit differentiation, and yields good approximate marginal posteriors. The Julia and Python MUSE packages\footnote{\url{https://cosmicmar.com/MuseInference.jl} and \url{https://cosmicmar.com/muse_inference}} have been updated to use implicit differentiation, and can solve problems defined by hand or with any of a number of popular probabilistic programming languages and automatic differentiation backends.
\end{abstract}


\section{Introduction}

MUSE is an algorithm for fast approximate hierarchical Bayesian inference, recently proposed by \cite{seljak2017,millea2022}. The user denotes some subset of model parameters as the ``parameters of interest," and the algorithm will approximate their marginal posterior while integrating out remaining ``latent'' parameters.   MUSE is efficient for very high-dimensional latent spaces and can often provide near-exact inference at orders of magnitude lower computational cost than other methods such as Hamiltonian Monte Carlo (HMC) or variational inference (VI) \cite{millea2022}.

The requirements for using MUSE on a given problem are that 1) samples can be generated from the prior and 2) gradients of the joint posterior probability distribution can be calculated. The latter requirement is the same as for HMC, VI, and many other tools. The former requirement is not strictly a requirement for some of these, but is generally even easier. All problems defined via a probabilistic programming language satisfy the requirements automatically. Owing to its reliance on prior samples, MUSE can be considered a form of simulation-based inference, extended to use readily available joint posterior gradients, similar to the proposal by \cite{cranmer2020}.

At its core, MUSE is based on an approximation to the marginal score formed from solutions to a series of optimization problems. As part of the algorithm, we must compute derivatives of these solutions, and, in this work, we improve MUSE by making use of implicit differentiation (ID) to perform this calculation. While ID is not a new development, it has recently been shown to be particularly powerful in conjunction with automatic differentiation (AD) \cite{duvenaud2022,blondel2022}. We follow this approach, and demonstrate that it leads to significant improvements in both speed and usability for MUSE, strengthening its case as a generic inference tool. Code to reproduce our results is available here\footnote{\url{https://github.com/marius311/muse-implicit-paper}}.

\vspace{-0.3cm}

\section{Summary of the MUSE method}

Here we give a brief and practical summary of MUSE to help understand where ID fits in (for a comprehensive introduction, see \cite{millea2022}). \MUSE is applicable to inference problems where the posterior probability of some parameters of interest, $\theta$, given data, $x$, requires marginalization over a high-dimensional latent space parameterized by $z$,
\begin{align}
    \mathcal{P}(\theta\,|\,x) = \int d^n\!z \, \mathcal{P}(x,z\,|\,\theta) \, \mathcal{P}(\theta).
\end{align}
The algorithm provides a fast estimate of the marginal posterior mean and covariance, which is computed under an approximation to the integral over $z$. This approximation involves solving a series of optimization problems wherein we maximize the joint likelihood, $ \mathcal{P}(x,z\,|\,\theta)$, over the latent parameters $z$, given fixed $x$ and $\theta$,
\begin{align}
    \label{eq:zmap}
    \hat z(\theta,x) \equiv \underset{z}{\rm argmax} \; \log \mathcal{P}(\x,z\,|\,\theta).
\end{align}
These correspond to maximum a posteriori (MAP) estimates of $z$, and they are used to define the score at the MAP,
\begin{align}
    \label{eq:smap}
    s_i^{\rm MAP}(\theta,\x) \equiv \frac{d}{d\theta_i}\log \mathcal{P}(\x,\hat z(\theta,x)\,|\,\theta).
\end{align}
The \MUSE estimate of the posterior mean, $\bar\theta$, is then implicitly defined as the solution to
\begin{align}
    s_i^{\rm MAP}(\bar\theta,\x) = \Big \langle s_i^{\rm MAP}(\bar\theta,\x) \Big \rangle_{x\sim\mathcal{P}(x\,|\,\bar\theta)},
    \label{eq:musesoln}
\end{align}
and the posterior covariance is $\Sigma\,{=}\,H^{-1} J H^{-\dagger}$, with
\begin{align}
    J_{ij} &= \Big\langle \smap[i](\bar\theta,\x) \, \smap[j](\bar\theta,\x) \Big\rangle_{\x\sim\mathcal{P}(\x\,|\,\bar\theta)}  -\Big\langle \smap[i](\bar\theta,\x) \Big\rangle \Big \langle \smap[j](\bar\theta,\x) \Big\rangle_{\x\sim\mathcal{P}(\x\,|\,\bar\theta)} \label{eq:J} \\
    H_{ij} &= \left. \frac{d}{d\theta_j} \left[ \Big\langle \smap[i](\bar\theta,\x) \Big\rangle_{\x\sim\mathcal{P}(\x\,|\,\theta)} \right] \right|_{\theta=\bar\theta}. \label{eq:H}
\end{align}

This definition gives \MUSE a number of useful properties (see \cite{millea2022} for proofs): 1) it is an asymptotically unbiased estimate of $\theta$ irregardless of any non-Gaussianity in the likelihood, 2) it is asymptotically optimal for a Gaussian likelihood, where it becomes equivalent to the marginal maximum likelihood estimate and the covariance becomes the inverse Fisher information matrix, 3) no dense operators of the dimensionality of $z$ ever need to be computed, meaning it is well-suited for high-dimensional problems and 4) it requires few tuning parameters, setting it apart from HMC, VI, or many other simulation-based inference methods, which need user-provided mass matrices, surrogate distributions, or neural network architectures to work or to achieve optimal performance on complicated latent spaces. MUSE is approximate, so it does not aim to generically replace exact algorithms like HMC, but in many cases, its speed and aforementioned properties make it a very advantageous alternative.

In practice, the optimization problem in Eq.~\eqref{eq:zmap} is performed with LBFGS using user-provided or AD gradients. An existing challenge for \MUSE is that naively computing Eq.~\eqref{eq:H} with AD would require propagating second-order derivatives through the optimizer, since a chain rule term involving $d\hat z/d\theta$ arises. With few or no AD libraries robustly supporting second-order AD through an optimizer, we have previously resorted to computing this term with finite differences (FD). This has not been completely prohibitive as FD are needed only over the low-dimensional $\theta$ and not over the high-dimensional $z$, so the solution remains tractable despite a linear computational scaling with the dimensionality of $\theta$. However, it requires tuning the FD step size for each dimension of $\theta$, and can at times incur large numerical errors. The main development of this paper is to demonstrate that this term can instead be computed more simply and exactly with ID.

\section{Using implicit differentiation}

To compute $H$ with ID, first note that Eq.~\eqref{eq:H} can be written as $H_{ij} = \frac{1}{N} \sum_{\alpha=1}^{N} h_{ij}(\Omega_\alpha)$, where $\Omega_\alpha$ are some independent random states, and
\begin{align}
    h_{ij}(\Omega) = \left. \frac{d}{d\theta^\prime_j}  \frac{d}{d\theta_i} \log \mathcal{P}\Big(x\big(\Omega,\theta^\prime\big), \hat{z}\big(x(\Omega,\theta^\prime),\bar\theta\big)\,\Big\vert\,\theta\Big) \right|_{\theta=\theta^\prime=\bar\theta}.
    \label{eq:h}
\end{align}

Here, we consider a single realization of $x$ as dependent on $\theta$ in the sense that any simulated $x$ can be written as a deterministic function of $\theta$ and a random state (think of $\Omega$ as the machine's pseudo random number generator). Expanding the chain rule once and omitting $\Omega$ and the final evaluation at $\bar\theta$ for brevity yields
\begin{align}
    \frac{d}{d\theta^\prime_j}  \frac{d}{d\theta_i} \log \mathcal{P}\Big(x\big(\theta^\prime\big), \hat{z}\big(x(\bar\theta),\bar\theta\big)\,\Big\vert\,\theta\Big) \; + 
    \left. \frac{d}{dz_n}  \frac{d}{d\theta_i} \log \mathcal{P}\Big(x\big(\bar\theta\big), z\,\Big\vert\,\theta\Big) \right|_{z=\hat z} \frac{d\hat{z}_n\big(x(\theta^\prime),\bar\theta\big)}{d\theta^\prime_j}.
    \label{eq:Him1}
\end{align}

The first term can be computed with second-order AD through the likelihood and through the prior samples of $x$. In practice, this means simply using the same random state on the forwards and/or backwards AD passes and otherwise considering random number generation constant (this is the default in most AD libraries). The second term, where we will use ID, involves a derivative of the MAP solution, $d\hat z/d\theta$. The MAP solution by definition obeys
\begin{align}
    \left. \frac{d}{dz} \log \mathcal{P}\Big(x\big(\theta^\prime\big), z \,\Big\vert\,\bar\theta\Big) \right|_{z=\hat{z}(x(\theta^\prime),\bar\theta)} = 0.
\end{align}
Taking a $\theta^\prime$ derivative of this equation and solving the resulting equation for $d\hat z/d\theta^\prime$ yields
\begin{align}
    \frac{d\hat z_n}{d\theta^\prime_j} = \left[ \frac{d^2}{dz_m dz_n} \log \mathcal{P}\Big(x\big(\bar\theta\big), z \,\Big\vert\,\bar\theta\Big) \right]^{-1} \left. \frac{d}{d\theta^\prime_j}\frac{d}{dz_m} \log \mathcal{P}\Big(x\big(\theta^\prime\big), z \,\Big\vert\,\bar\theta\Big) \right|_{z=\hat z}.
    \label{eq:Him2}
\end{align}
This quantity now only requires derivatives through the likelihood rather than through an optimizer; in fact, it is independent of the particular optimizer used to obtain $\hat z$. Computing it involves solving a linear problem with the same dimensionality as $z$. Because $z$ is assumed high-dimensional where forming an explicit matrix is impossible, we solve the system iteratively, with the action of the quantity in brackets above given by a jacobian-vector product. Note, however, that the linear operator is symmetric since it is a Hessian, and, by definition if the MAP exists (which is a requirement for MUSE anyway), it is positive definite. Thus, we can use an efficient conjugate gradient solver which exploits this structure, as opposed to generic linear solvers which must be used in more general ID problems. 

\section{Results}
\label{sec:results}

We compare HMC and MUSE with or without ID on three representative inference problems:

{\bf Funnel problem} \quad We consider an embedding of several Neal's funnels into a toy hierarchical problem \cite{neal2003,millea2022}. The model is:
\begin{align}
    \theta_i \sim {\rm Normal}(0, 3) \quad\quad 
    z_{ij} \sim {\rm Normal}(0, \exp(\theta_i/2)) \quad\quad
    x_{ij} \sim {\rm Normal}(\tanh(z_{ij}), 1)
\end{align}
with $i\,{\in}\,1{:}10$ parameters and $j\,{\in}\,1{:}500$ latent dimensions per parameter. Although the embedded funnels are independent, for demonstration, we solve the entire problem as one large system when running either NUTS or MUSE. We note that MUSE was originally developed for cosmological applications, and this problem is conceptually extremely close to typical field-level cosmological inference problems: the $\theta$ serve the role of ``power-spectrum'' amplitudes, the $z$ serve as some initial Gaussian random field, and the hyperbolic tangent serves as some non-linear field evolution.

{\bf Bayesian Neural Network} \quad Following the example given in \cite{numpyro-bnn}, we consider a Bayesian neural network (BNN) analysis, where we interpolate some noisy one-dimensional data with a three-layer neural network. The model is:
\begin{align}
    \begin{split}
        \sigma_i &\sim {\rm LogNormal}(0,1) \\
        \tau &\sim {\rm Gamma}(3,1)
    \end{split}
    \begin{split}
        W_i &\sim {\rm Normal}(0,\sigma_i) \\
        Y_j &\sim {\rm Normal}({\rm NN}(W_i), 1/\tau)
    \end{split}
\end{align}
where $i\,{\in}\,1{:}3$ layers, the layer weights, $W_i$, which parameterize the network, NN, contain 45 latent dimensions and map the data coordinates to 5 hidden units and finally to the data space, and the data $Y_j$ consists of $j\,{\in}\,1{:}500$ data points. The goal is to infer the $\sigma_i$ and $\tau$. We note that each internal optimization solution in Eq.~\eqref{eq:zmap} involves training the network given some prior on the weights. For this simple example we use our standard LBFGS solver, but other more machine learning oriented solvers can readily be used for the internal MUSE optimization step as well.

{\bf Probabilistic Principal Component Analysis} \quad Finally, we consider a probabilistic principal component analysis (PPCA) with automatic relevance determination \cite{bishop2006}. The model is:
\begin{align}
    \begin{split}
        \alpha_i &\sim {\rm InverseGamma}(1, 1) \\
        Z_{ij} &\sim {\rm Normal}(0, 1)
    \end{split}
    \begin{split}
        W_{ki} &\sim {\rm Normal}(0, \sqrt\alpha_i) \\
        X_{kjl} &\sim {\rm Normal}(W_{ki} Z_{ij}, 1)
    \end{split}
\end{align}
with $i\,{\in}\,1{:}10$ principal components, $j,k\,{\in}\,1{:}100$ observations, and $l\,{\in}\,5$ batches. The goal is to find the largest principal component amplitudes, $\alpha$, given observations of $X$, while marginalizing over the entries in the $Z$ and $W$ matrices.

\begin{figure}
    \centering
    \includegraphics[width=0.45\textwidth,valign=t]{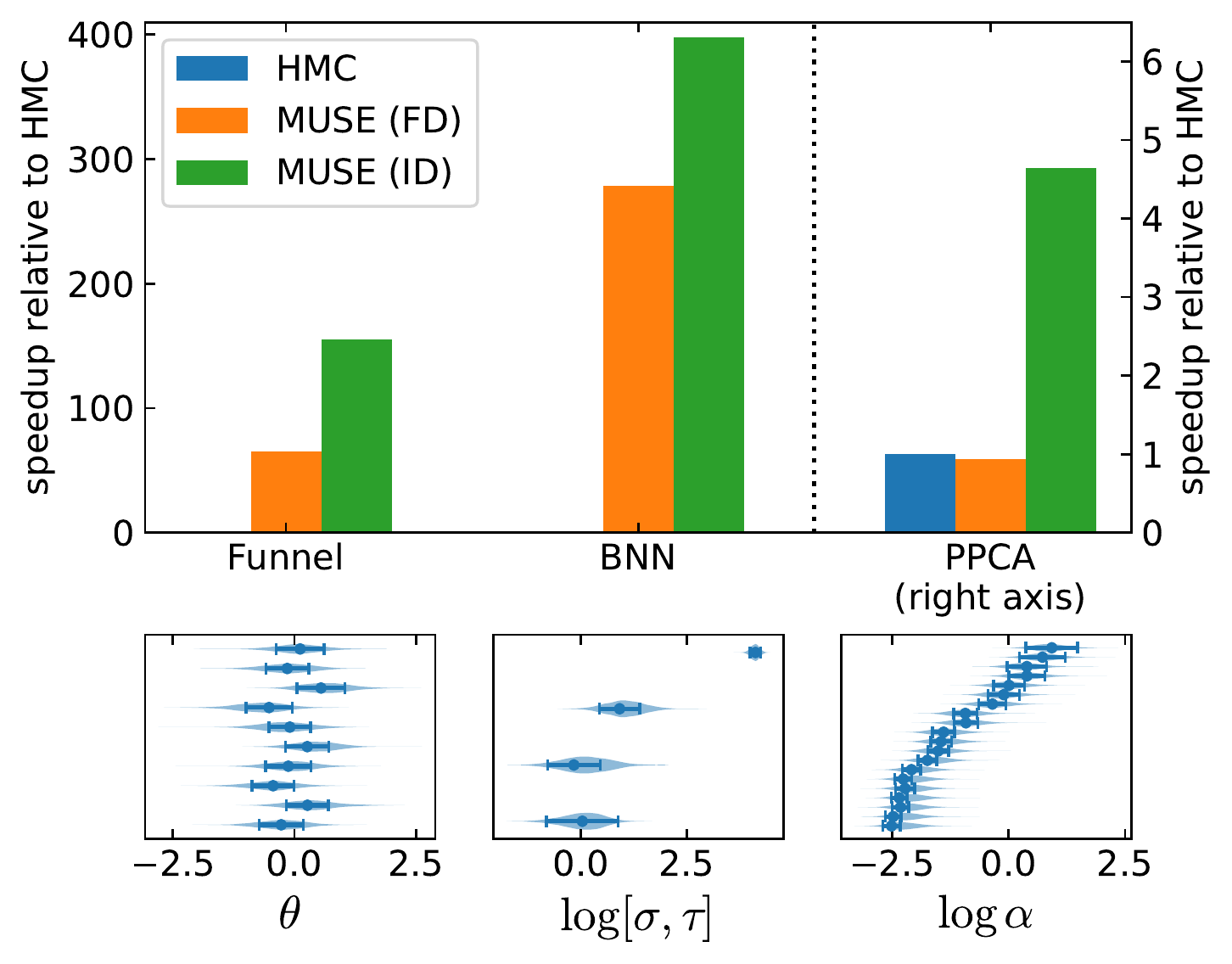}
    \includegraphics[width=0.45\textwidth,valign=t]{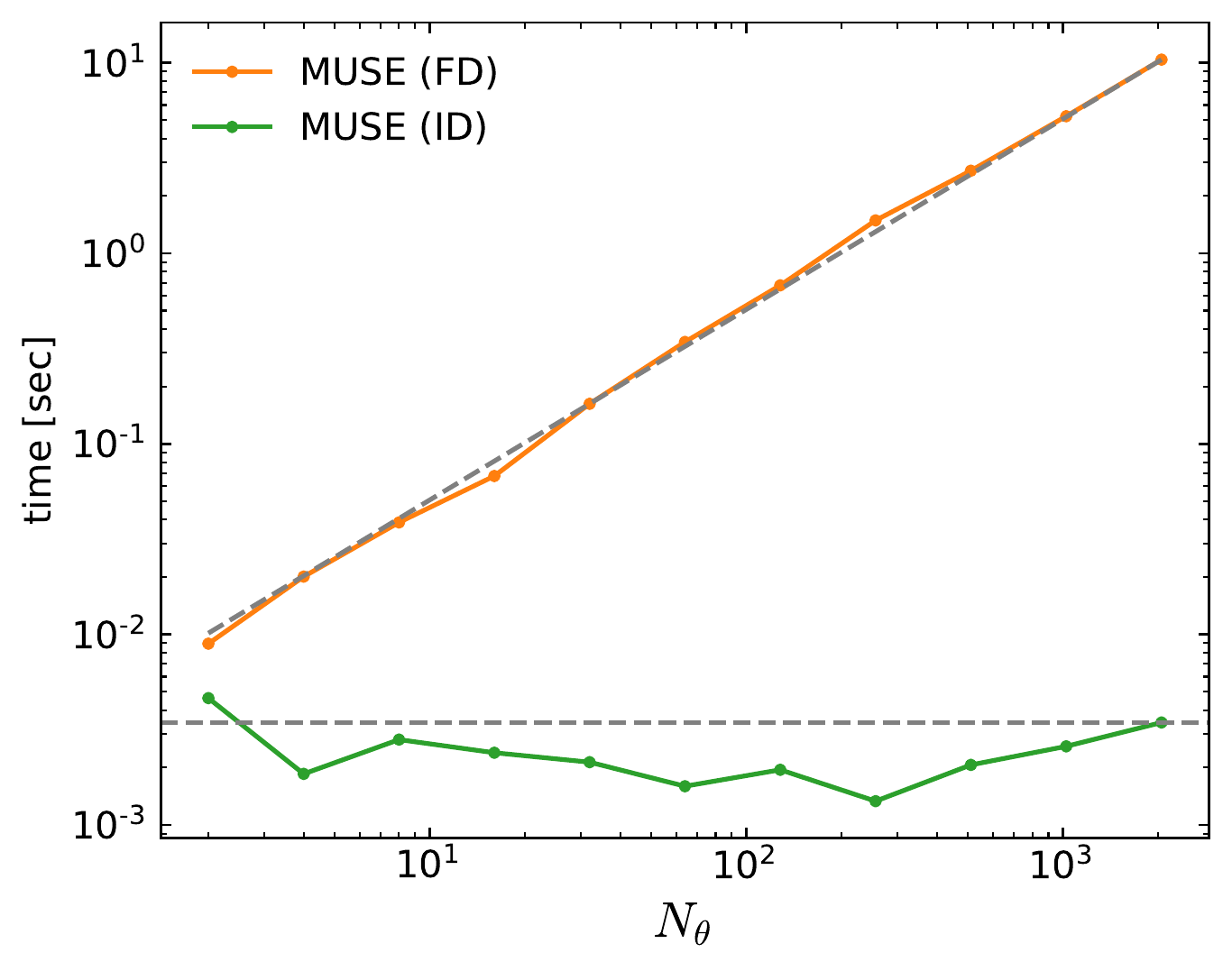}
    \caption{{\it (Top left)} Speedups which are possible with MUSE both with or without ID as compared to HMC on a variety of hierarchical Bayesian inference problems (described in Sec.~\ref{sec:results}). {\it (Bottom left)} HMC posteriors as violin plots, compared to MUSE results as error bars. {\it (Right)} Empirical check of the asymptotical scaling of the $H$ computation with FD or ID.}
    \label{fig:fig1}
\end{figure}

Our benchmarks compare the number of posterior gradient evaluations needed such that for all parameters of interest, we reach 1) a 10\% error on the mean relative to the standard deviation and 2) a 10\% relative error on the standard deviation. Given a Gaussian sampling distribution, these criteria impose the same constraint. For HMC, this corresponds to achieving an effective sample size of 100 for all parameters. We use NumPyro to implement each model \cite{phan2019,bingham2019}, and sample with NumPyro's NUTS implementation with default parameters. For MUSE this corresponds to running MUSE with 100 simulations and setting the $\theta$ tolerance to 10\%. We use the existing Jax \cite{jax2018github} MUSE implementation to run MUSE on the same NumPyro model.

The results are summarized in the left panels in Fig.~\ref{fig:fig1}. We see that for each of the three problems, ID outperforms the previous FD approach by as much as a factor of 5. In all cases, MUSE with ID significantly outperforms HMC, including by a factor of 391 in the most dramatic case (the BNN). The bottom panel shows a comparison of the inferred values of the parameters of interest, confirming the quality of the MUSE approximation.

We also expect a more favorable computational scaling for ID over FD as we increase the dimensionality of $\theta$. This is because computing $h_{ij}$ with FD requires perturbing each element of $\theta$ and recomputing a MAP each time, whereas ID requires just one MAP that is then used evaluating all terms in Eqn.~\eqref{eq:Him1}, with the tradeoff of also needing to solve a linear problem. To confirm this tradeoff is beneficial, we modify our funnel problem to increase the dimensionality of $\theta$ (while keeping the latent dimensionality the same), and plot resulting timings for the $H$ computation in the right panel of Fig.~\ref{fig:fig1}. We find that FD scales linearly with the dimensionality of $\theta$ as expected, but that ID is nearly constant, meaning the cost of the linear solver is subdominant. For the configurations considered, we reach multiple orders of magnitude speedups over FD.

\section{Conclusions}

In this work, we have shown that ID makes the MUSE algorithm faster and removes reliance on numerically-noisy FD. It requires second-order derivatives through the joint likelihood, but not through any optimizer, and never fully with respect to the latent space, meaning MUSE with ID is still well-suited for very high-dimensional problems. We have also provided examples of MUSE applied to BNNs and PPCA, demonstrating the extended applicability of the algorithm, which had previously been tested only on simpler toy problems or more complex but less general problems in cosmology \cite{horowitz2019,millea2022}. Beyond speed and accuracy improvements, removing the need to verify or tweak FD step-sizes represents a significant usability advancement for the algorithm.

\begin{ack}
I would like to thank Guillaume Dalle for his \hyperlink{https://www.youtube.com/watch?v=TkVDcujVNJ4}{JuliaCon 2022} talk on implicit differentiation which inspired this work, and Andreas Noack for a helpful discussion on automatic differentiation in the presence of random number generators. This work was partially supported by the National Science Foundation through grants OPP-1852617, 1814370, and 1839217. This research used resources of the National Energy Research Scientific Computing Center (NERSC), a DOE Office of Science User Facility supported by the Office of Science of the U.S. Department of Energy under Contract No. DE-AC02-05CH11231.
\end{ack}

\bibliographystyle{unsrt}
\bibliography{marius}

\end{document}